%% file: main.tex
\documentclass[10pt,english]{article}
\usepackage[T1]{fontenc}

\usepackage[margin=0.9in]{geometry}

\usepackage{babel}
\usepackage[utf8]{inputenc} 
\usepackage[T1]{fontenc}    
\usepackage{hyperref}       
\usepackage{url}            
\usepackage{booktabs}       
\usepackage{amsfonts}       
\usepackage{nicefrac}       
\usepackage{microtype}      
\usepackage[dvipsnames]{xcolor}
\usepackage{amsmath,graphicx}
\usepackage{algorithm}
\usepackage[noend]{algpseudocode}
\usepackage{amsmath,amssymb}
\usepackage{subcaption}
\DeclareMathOperator*{\argmin}{arg\,min}

\DeclareUnicodeCharacter{2212}{-}
\usepackage{textcomp}

\let\bm\boldsymbol

\title{Sparse Generative Adversarial Network}
\author{
  Shahin Mahdizadehaghdam\\
  Department of Electrical\\ and Computer Engineering\\
  North Carolina State University\\
  Raleigh, NC 27695 \\
  \texttt{smahdiz@ncsu.edu}
  \and
  Ashkan Panahi\\
  Department of Electrical\\ and Computer Engineering\\
  North Carolina State University\\
  Raleigh, NC 27695 \\
  \texttt{apanahi@ncsu.edu}
  \and
  Hamid Krim\\
  Department of Electrical\\ and Computer Engineering\\
  North Carolina State University\\
  Raleigh, NC 27695 \\
  \texttt{ahk@ncsu.edu}
}

\begin{document}

\maketitle

\input{abstract.tex}

\input{introduction.tex}
\input{Related.tex}
\input{Basis.tex}

\input{ProposedMethod.tex}
\input{Results.tex}

\input{conclusions.tex}

{\small
\bibliographystyle{ieee}
\bibliography{refs}
}

\end{document}

%% file: abstract.tex
\begin{abstract}
\label{sec:abstract}
We propose a new approach to Generative Adversarial Networks (GANs) to achieve an improved performance with additional robustness to its so-called and well recognized mode collapse.
We first proceed by mapping the desired data onto a frame-based space for a sparse representation to lift any limitation of small support features prior to learning the structure. To that end  we start by
dividing an image into multiple patches and modifying the role of the generative network from producing an entire image, at once, to creating a sparse representation vector for each image patch. We synthesize an entire image by multiplying generated sparse representations to a pre-trained dictionary and assembling the resulting patches.
This approach restricts the output of the generator to a particular structure, obtained by imposing a Union of Subspaces (UoS) model to the original training data, leading to more realistic images, while maintaining a desired diversity. To further regularize GANs in generating high-quality images and to avoid the notorious mode-collapse problem, we introduce a third player in GANs, called reconstructor. This player utilizes an auto-encoding scheme to ensure that first, the input-output relation in the generator is injective and second each real image corresponds to some input noise. We present a number of experiments, where the proposed algorithm shows a remarkably higher inception score compared to the equivalent conventional GANs.
\end{abstract}

%% file: introduction.tex
\section{Introduction}
\label{sec:intro}
In recent years, Generative Adversarial Networks (GANs)  \cite{radford2015unsupervised, karras2017progressive, zhang2017stackgan++} showed impressive results in various image generation problems, such as image
super-resolution \cite{ledig2017photo, denton2015deep, tran2019nonlinear},
dialogue generation \cite{li2017adversarial},
and image translation \cite{zhu2017toward, yi2017dualgan, liu2017unsupervised, isola2017image}.
As an implicit method of probability density estimation, GANs
commonly consist of two main units: A generator function that synthesizes various images from different realizations of an input noise vector, and a discriminator function which examines the quality of the produced images by the generator.
The generator and discriminator are usually differentiable-functions based on deep convolutional networks.
In GANs, the input noise vector is transformed by the generator to produce a "fake" image.  The discriminator function receives either a fake sample from the generator or an "original" sample from the data set and decides on its genuineness. The two functions are successively trained in a min-max optimization framework \cite{goodfellow2014generative}, where the weights of the generator network are gradually adapted to generate more realistic images, in order to fool the discriminator.

The research area of GANs still takes its first steps toward full growth. Due to the complex structure of real-world images, the images generated by GANs widely lack details and do not usually look completely realistic. Images are known to follow complicated probability distributions, and estimating these distributions at the full image scale is generally a difficult task. Mode collapse, a well-known phenomenon in GANs, where highly similar images are frequently generated from different inputs is widely attributed to the problem of complexity of the image distributions (small support events). 
To alleviate the above-mentioned issues, we propose a novel generator model which learns the structure of images in smaller scales and uses the learned structures to synthesize full-scale images. To this end, we adopt dictionary learning and sparse representation \cite{mairal2009online}, as an effective method to learn and leverage the structure of data.
Parsimonious data representation by learning overcomplete dictionaries has shown promising results in a variety of problems such as image denoising \cite{elad2006image}, image restoration \cite{xu2016cloud}, audio processing \cite{grosse2012shift}, and image classification \cite{Zhang2016168}.  
The learned dictionaries constitute a frame, whose atoms are employed to linearly represent data vectors with their corresponding sparse vectors of coefficients \cite{wright2009robust}. (This effectively lifts the singular support \cite{guo_paper}).

We hence propose a novel generator network which generates a sparse vector of coefficients for each image patch instead of generating the entire image. 
Full-size output images are assembled by tiling the image patches, produced by multiplying the generated sparse coefficient vectors to a pre-trained dictionary.
This approach avoids the complexities with the conventional method of generating the entire image in one step. Instead, it affords a multi-stage solution, where at the first stage, simpler structures at the scale of image patches are generated, and a full-size image is assembled at the final stage. Incorporating sparse representations also limits the search space of the produced images to a linear combination of a set of dictionary atoms, tailored for representing real-world images, enhancing the procedure of generating realistic images.
To further regularize with respect to  the mode collapse problem, we also introduce a third player to GANs that we call the reconstructor. 
The reconstructor has two goals. First, it makes sure that different inputs of the generator result in dissimilar outputs i.e., the generator is an injective map. Second, it guarantees that the real images in the dataset can also be synthesized from some input noise i.e., the range space of the generator includes the entire set of real images. These goals are  obtained by treating the generator network as a "decoder" from a latent input space and simultaneously training an "encoder" network, which reversely computes the input of the generator from its output. The encoder function as a regularizer to the generator network is minimized successively along with the generator and the discriminator losses. Existence of such an auto-encoder scheme guarantees that the generator network is injective and model collapse is avoided. Moreover by training this auto-encoder scheme on real images, we achieve the second goal. 

The balance of this paper is organized as follows: 
In Section \ref{sec:Related}, we provide an overview to the state of-the-art works relevant to this paper.
In Section \ref{sec:Basis}, we briefly recall the mathematical basis of GANs, as well as some  background information of relevance to this paper. 
We formulate and propose our new approach in Section \ref{sec:model}.  
Substantiating experimental results are presented in Section \ref{sec:Results}. Finally, we provide some concluding remarks in Section \ref{sec:conclusions}.



%% file: Related.tex
\section{Related Studies}
\label{sec:Related}
There have been plenty of different studies on GANs since their advent in \cite{goodfellow2014generative}. 
We summarize these works into three main groups and briefly overview them.

\textit{Stable training of GANs}: 
Due to the non-convex nature of the underlying optimization problems in GANs, they are notoriously difficult to train. In particular, they are generally prone to non-convergence, diminishing gradient and mode collapse, the latter happening when the generator frequently outputs a narrow set of highly similar samples. Several efforts were devoted to alleviating these problems. An autoencoder-based regularization was proposed in \cite{che2016mode} by penalizing missing modes.
To further mitigate the mode collapse problem, an unrolled optimization of the discriminator is proposed in \cite{metz2016unrolled}. This technique results in a more power-balanced generator and discriminator and is successful in preventing mode collapse, but the computational cost is high.
In WGAN \cite{arjovsky2017wasserstein}, the Earth-Mover (Wasserstein-1) distance is adopted as the objective for the generator. 
This objective is approximated by restricting the discriminator to 1-Lipschitz function through weight clipping.
An enhancement over WGAN is proposed in \cite{gulrajani2017improved}. In this work, a different method for bounding the gradients is proposed by adding a gradient penalty term to the objective.
Furthermore, in a recent effort \cite{miyato2018spectral}, the Lipschitz constant of the discriminator function is controlled by limiting the spectral norm of the weights in the discriminator.

\textit{Architectures of GANs}: 
The Deep Convolutional Generative Adversarial Network, DCGAN, architecture is proposed in \cite{radford2015unsupervised}. This architecture is usually a set of (four) fractionally-strided convolutional layers with no pooling and fully connected layers.
Stacked Generative Adversarial Networks, StackGAN \cite{zhang2017stackgan}, aims to generate 256 $\times$ 256 realistic images from text descriptions in two steps. Initially, lower resolution images are generated from text descriptions, and then, higher-resolution images are generated from low-resolution images.
In \cite{karras2017progressive}, the authors proposed to progressively develop the generator and discriminator as training continues. This method helps to speed up the training and improve image quality.

\textit{Applications of GANs}: GANs have been utilized in different areas of machine learning, including Natural language processing (NLP) and computer vision.
In \cite{zhang2016generating} an LSTM network and a CNN have been trained in an adversarial way to generate realistic text. The LSTM network takes the random input vector and generates text while the CNN discriminates between real text and generated one.
An unsupervised word translation method has been proposed in \cite{conneau2017word}, which translates words without any cross-lingual supervision. In this work, a generator network has been used to map word embeddings from one domain to another, while the discriminator aims to detect the origin of the embedding.
A variation of GANs with class label information, known as Conditional GANs, have been utilized in image-to-image translation \cite{isola2017image} and demonstrated successful performance in reconstructing objects from edge maps and colorizing tasks.
The generator network is used in \cite{ledig2017photo, sonderby2016amortised} to produce super-resolution images from low resolution images. The discriminator networks in these works are trained to discriminate between super-resolved images and real high-resolution images.
GANs are also used in transforming person images to arbitrary poses and synthesizing clothing images and styles from an image \cite{yoo2016pixel, ma2017pose}.

From architectural point of view, our proposed approach is similar to DCGAN \cite{radford2015unsupervised} with an additional provision for a layer to generate sparse coefficients. We use the WGAN technique in \cite{gulrajani2017improved} to stabilize the training of GANs.
Furthermore, we propose a third player to GANs (reconstructor) which assists the generator network in regularizing and generating more realistic images. The reconstructor network in GANs, to the best of our knowledge, is a novel idea without precedent.

%% file: Basis.tex
\section{Mathematical Basis of GANs}
\label{sec:Basis}
The initial formulation of GANs in \cite{goodfellow2014generative} is as follows,
\begin{equation}
\begin{aligned}
\min_{G} \max_{D} ~ &\mathbb{E}_{\bm{x}\sim p_{\bm{x}}} [\log~ D(\bm{x})]+
 &\mathbb{E}_{\bm{z}\sim p_{\bm{z}}} [\log (1- D(G(\bm{z})))],
\end{aligned}
\label{eq:GAN_goodfellow}
\end{equation}
where $\bm{x}$ and $\bm{z}$ are the real data and the noise input vectors respectively and $D$ and $G$ are the discriminator and the generator networks. $p_{\bm{x}}$ and $p_{\bm{z}}$ are the distributions of the real data and noise vector respectively. 

The above optimization problem is solved by an alternating optimization scheme. At each step, the generator or the discriminator variables are fixed and the problem is solved for the other player. 
Defining $p_{\bm{g}}$ as the distribution of the generated images ($G(\bm{z}) \sim p_{\bm{g}}$), Eqn. (\ref{eq:GAN_goodfellow}) is equivalent to minimizing the Jensen-Shannon divergence between  $p_{\bm{x}}$ and $p_{\bm{g}}$ \cite{goodfellow2014generative}. 

The Jensen-Shannon divergence between $p_{\bm{x}}$ and $p_{\bm{g}}$ is not always continuous with respect to the variables in the generator \cite{arjovsky2017wasserstein} and this leads to an instability in training GANs. Therefore, in \cite{arjovsky2017wasserstein}, the Jensen-Shannon divergence is replaced by the earth-mover's distance $W(p_{\bm{x}}, p_{\bm{g}})$ which is continuous everywhere and leads to the following optimization problem,
\begin{equation}
\begin{aligned}
\min_{G} \max_{D \in \mathcal{D}} ~ \mathbb{E}_{\bm{x}\sim p_{\bm{x}}} [D(\bm{x})] - 
 \mathbb{E}_{\bm{z}\sim p_{\bm{z}}} [ D(G(\bm{z})) ],
\end{aligned}
\label{eq:WGAN}
\end{equation}
where $\mathcal{D}$ is the set of 1-Lipschitz functions. 
Considering an optimal discriminator function, the solution of the above optimization problem minimizes the earth-mover's distance between $p_{\bm{x}}$ and $p_{\bm{g}}$. 

In order to impose the Lipschitz constraint on the discriminator ($D \in \mathcal{D}$), in \cite{arjovsky2017wasserstein}, the weights of the discriminator are simply clipped within a compact interval $[−c, c]$. Alternatively, the following gradient penalty term is minimized in \cite{gulrajani2017improved} to impose the Lipschitz constraint.
\begin{equation}
\begin{aligned}
\mathbb{E}_{\hat{\bm{x}}\sim p_{\hat{\bm{x}}}} [ (||\nabla_{\hat{\bm{x}}} D(\hat{\bm{x}} )||_2 - 1)^2  ],
\end{aligned}
\label{eq:ImWGAN}
\end{equation}
where $\hat{\bm{x}} \sim p_{\hat{\bm{x}}}$ is a random sample, uniformly randomly chosen on the straight line segment connecting $\bm{x}$ and $G(\bm{z})$. The gradient penalty in Eqn. (\ref{eq:ImWGAN}) is motivated from the fact that 
a function is 1-Lipschtiz if and only if its gradient norm is upper bounded by 1. Authors in \cite{gulrajani2017improved} showed that the gradient penalty in Eqn. (\ref{eq:ImWGAN}) is more effective in stabilizing training of GANs than the weight clipping technique.

%% file: ProposedMethod.tex
\section{The Proposed Method}
\label{sec:model}
In contrast to Eqn. (\ref{eq:GAN_goodfellow}), our proposed formulation of GAN is composed of three players: a discriminator function which decides whether an input is a real or a generated sample, a generator which synthesizes images by initially generating sparse representations of image patches, and a reconstructor which ensures that the generator is capable of generating all the real samples from some associated noise signals.
We formulate our proposed method by defining the following optimization problem,
\begin{eqnarray}
\begin{aligned}
\min\limits_{G} \max\limits_{D \in \mathcal{D}} ~ \mathbb{E}_{\bm{x}\sim p_{\bm{x}}} [D(\bm{x})] - & 
\mathbb{E}_{\bm{z}\sim p_{\bm{z}}} [ D(G(\bm{z})) ]\\
 + \mathbb{E}_{\bm{x}\sim p_{\bm{x}}} & [||\bm{x} - G({E^*}(\bm{x}))||^2_2]\nonumber\\
\mathrm{s.t:} ~~~~~~~~~ E^*\in\arg\min\limits_{E}\mathbb{E}_{\bm{z}\sim p_{\bm{z}}} &[\|\bm{z}-E(G(\bm{z}))\|_2^2],
\end{aligned}
\label{eq:SP_GAN0}
\end{eqnarray}
where $D, G$ and $E$ are the discriminator, the sparse generator and the encoder networks, respectively. 
Moreover, $E^*$ is an optimal encoder network for $G$, which serves as a left inverse of $G$, according to the constraint. 
The discriminator belongs to the set $\mathcal{D}$ of $1-$Lipschitz functions and we explain the details of the Sparse Generator Network (SPGAN) in Section \ref{subsec:proposed_gan}.
$\bm{x}$ and $\bm{z}$ are the real data and the noise input vectors, respectively sampled from the real data distribution $p_{\bm{x}}$ and a fixed noise distribution $p_{\bm{z}}$.
The last term in Eqn. (\ref{eq:SP_GAN0}), minimizes the error between the input real image  $\bm{x}$ and the generated image by $G$ from the corresponding noise $z_{\bm{x}}= E^*(\bm{x})$ to the data point $\bm{x}$. This term will be more thoroughly discussed in Section \ref{subsec:proposed_rec}.

In order to solve the above optimization problem, we define the following three loss functions,
\begin{equation}
\begin{aligned}
& L_{D} = \mathbb{E}_{\bm{z}\sim p_{\bm{z}}} [ D({G}(\bm{z}))] -   \mathbb{E}_{\bm{x}\sim p_{\bm{x}}} [D(\bm{x})]+\\
& ~~~~~~~~~~ \lambda \mathbb{E}_{\hat{\bm{x}}\sim p_{\hat{\bm{x}}}} [ (||\nabla_{\hat{\bm{x}}} D(\hat{\bm{x}} )||_2 - 1)^2  ],\\
& L_{G} = - \mathbb{E}_{\bm{z}\sim p_{\bm{z}}}  [ D({G}(\bm{z}))],\\
& L_{R} = \mathbb{E}_{\bm{x}\sim p_{\bm{x}}} [||\bm{x} - G({E^*}(\bm{x}))||^2_2],
\end{aligned}
\label{eq:SP_GAN1}
\end{equation}
where $L_{D}$, $L_{G}$ and $L_{R}$ are respectively the loss functions for the discriminator, the generator and the reconstructor. We explain the details of the reconstructor loss $L_R$ in Section \ref{subsec:proposed_rec}. 
The formulation of $L_{D}$ is similar to the one in \cite{gulrajani2017improved} with the last term being the gradient penalty to enforce the Lipschitz constraint.
$\hat{\bm{x}} \sim p_{\hat{\bm{x}}}$ is a random sample, randomly uniform chosen on the straight line segment connecting $\bm{x}$ and $G(\bm{z})$. In the following, we explain the sparse generator network and the reconstructor loss in detail. 

\subsection{Sparse Generator Network}
\label{subsec:proposed_gan}
\begin{figure*}[!t]
    \centering
    \begin{minipage}[t]{0.48\textwidth}
        \centering
        \includegraphics[scale=0.34]{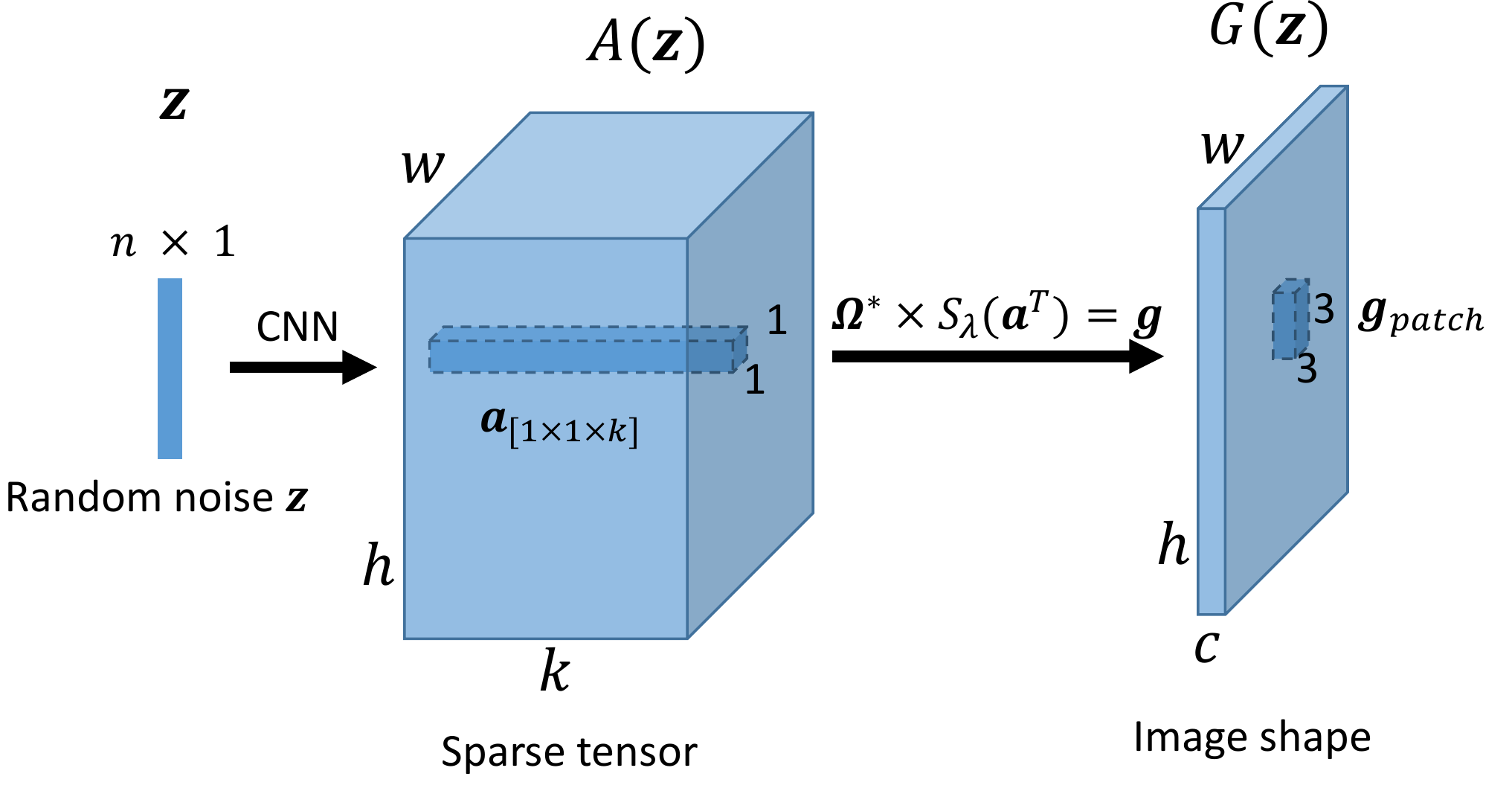}
        \captionof{figure}{Sequential steps of a sparse generator network.}
        \label{fig:SpGANS_arch_a}
    \end{minipage}~~
    \begin{minipage}[t]{0.43\textwidth}
        \centering
        \includegraphics[scale=0.48]{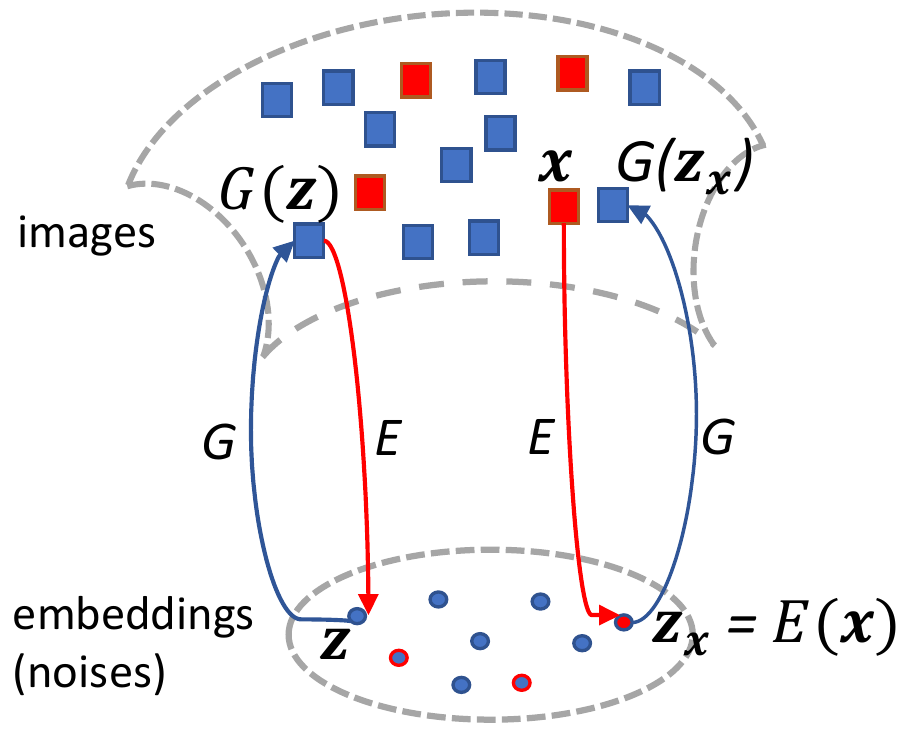}
        \captionof{figure}{Generating and encoding images. Blue/red squares and circles represent generated/real images and their associated latent vectors respectively.}
        \label{fig:SpGANS_arch_b}
    \end{minipage}
	\label{fig:SpGANS_arch}
\end{figure*}

As we elaborated in Section \ref{sec:intro}, our generator network consists of a deep neural network that generates a vector of sparse coefficients $S(\bm{z})$ from a random input noise vector $\bm{z}$ as an input. These coefficients are used as a representation of the image patches in a linear model. Hence, the image patches can be simply computed by multiplying the coefficients to a pre-trained dictionary $\bm{\Omega}$. Later, the generated patches are connected to form a full-size image. One can uniformly extract image patches from the training images. These image patches are generally in small sizes ($3 \times 3$) and may overlap with each other. Given vectorized image patches, $\bm{g}_i \in R^m \; i\in \{1,.., s\}$, as columns of a matrix $\bm{G}$, 
the dictionary $\bm{\Omega}^* \in R^{m\times k}$  is  trained by minimizing the following reconstruction loss,
\begin{equation}
\begin{aligned}
\{\bm{\Omega}^*, ~\bm{R}~ \} = \argmin_{\substack{\bm{\Omega},\;\bm{R} }}\; 
  \frac{1}{2}||\bm{G} - \bm{\Omega}\bm{R}||_F^2  + \lambda ||\bm{R}||_1,
\bm{\Omega} \in \mathcal{C},\\
\end{aligned}
\label{eq:dict_learning}
\end{equation}
where  the columns $\bm{r}_i \in R^k$ of the matrix $\bm{R}$ are the sparse representations of the image patches and $\mathcal{C}$ is a convex set of matrices with unit $L_2$-norm columns. 

In Fig. \ref{fig:SpGANS_arch_a}, we show the sequence of computational steps of generating an image from a random input noise vector and their associated specifications.
The random noise vector $\bm{z}$ is transformed to the tensor $A(\bm{z})$ of shape $w \times h \times k$. This transformation is by reshaping the input noise vector and applying transposed convolutions \cite{radford2015unsupervised}. 
The soft thresholding function  $S_{\lambda}(\bm{.})$ \cite{donoho1995noising} is applied to each depth-vector $\bm{a}^T$ to create a sparse vector. The sparse vector $S_{\lambda}(\bm{a}^T)$ multiplicatively scales a pre-trained dictionary $\bm{\Omega}^*$ to generate an image patch $\bm{g}$. The image patches are aligned next to each other (in overlapping areas we average the pixel values) to produce the full image $G(\bm{z})$.

In our method, the image patches are produced by generating the sparse vectors and multiplying them to the dictionary $\bm{\Omega}^*$. The generated image patches are therefore, limited to a linear combination of a small subset of the dictionary atoms. This constraint helps to ensure that besides generating new samples, the generator network can generate images that resemble the training samples. Note that
the image patches have smaller sizes and are less complicated in structure. 

\algrenewcommand\algorithmicindent{1.0em}%

\begin{algorithm}[!b]
\caption{}\label{alg}
\textbf{Initialization:}
\begin{algorithmic}[1]
\State{Initialize the discriminator, the generator and the encoder parameters ($w$, $\theta$, and $\phi$) randomly.}
\State{Train the dictionary $\bm{\Omega}$ by solving Eqn. (\ref{eq:dict_learning}})
\algstore{myalg}
\end{algorithmic}
\textbf{Training:}
\begin{algorithmic}[1]
\algrestore{myalg}
\While{$\theta$ has not converged}:

    \For{$t \text{ \textbf{in} } \{1,n_{discr.}\}$}:
        \For{$i \text{ \textbf{in} } \{1,b\}$}:
            \State{Sample $\bm{x}_i$, $\bm{z}_i$, and $\bm{\epsilon}_i$ from $p_{\bm{x}}$, $p_{\bm{z}}$, and the uniform distr. $U[0,1]$ respectively,}
            \State{$ \hat{\bm{x}}_i \leftarrow \bm{\epsilon}_i \bm{x}_i + (1-\bm{\epsilon}_i)  G_\theta(\bm{z}_i)$}
            \State{$L^i_{D} \leftarrow  D_w(G_\theta(\bm{z}_i)) -  D_w(\bm{x}_i) + \lambda  (||\nabla_{\hat{\bm{x}}} D_w(\hat{\bm{x}}_i )||_2 - 1)^2$},
            \State{$w \leftarrow w - \nabla_w (\frac{1}{b} \sum_{i=1}^b L^i_{D})$},
        \EndFor
    \EndFor

    \For{$i \text{ \textbf{in} } \{1,b\}$}:
        \State{Sample latent variable $\bm{z}_i$, from $p_{\bm{z}}$,}
        \State{$L^i_{G} \leftarrow  - D_w(G_\theta(\bm{z}_i))$,}
        \State{$\theta \leftarrow \theta - \nabla_{\theta} (\frac{1}{b} \sum_{i=1}^b L^i_{G})$},
    \EndFor

    \For{$t \text{ \textbf{in} } \{1,n_{reconst.}\}$}:
        \State{Train the encoder network Eqn. (\ref{eq:encoder})},
        \For{$i \text{ \textbf{in} } \{1,b\}$}:
            \State{Sample real data $\bm{x}_i$ from $p_{\bm{x}}$. Encode the real image to latent variable $\bm{z}_{\bm{x}_i}=E^*(\bm{x_i})$ },
            \State{$L^i_{R} \leftarrow ||\bm{x}_i - {G}(\bm{z}_{\bm{x}_i})||_2 $,}
            \State{$\theta \leftarrow \theta - \nabla_{\theta} (\frac{1}{b} \sum_{i=1}^b L^i_{R})$,}
        \EndFor
    \EndFor

\EndWhile
\end{algorithmic}
\end{algorithm}

\subsection{Reconstructor Loss}
\label{subsec:proposed_rec}
In this section, we explain the details of the reconstructor loss $L_{R}$ in Eqn. (\ref{eq:SP_GAN1}). Considering the noise vectors as samples from a latent space, $G(\bm{z})$ can be interpreted as a map (decoder) from the latent space to the space of generated images (blue arrows in Fig. \ref{fig:SpGANS_arch_b}). 
The role of the generator network is to produce an image for any sample from the latent space. We also ensure that the map $G(\bm{z})$ can generate a wide variety of real-world images by verifying that it is injective and hence mode-collapse is hard. To this end, we require an "encoder" function, $E$, to exist such that $E(G(\bm{z}))=\bm{z}$ approximately holds true for every $\bm{z}$. Figure \ref{fig:SpGANS_arch_b} explains why this requirement guarantees injectivity: If two realizations $\bm{z}_1,\bm{z}_2$ are mapped to the same image $\bm{x}$, then it is impossible for $E$ to map $\bm{x}$ back to the latent space.
We obtain the encoder $E$ by employing a neural network $E_\phi$ and performing the optimization in the constraint of Eq.~\eqref{eq:SP_GAN0} by solving the following optimization problem 
with respect to the parameters of the encoder network, $\bm{\phi}$,
\begin{equation}
\begin{aligned}
\bm{\phi}^*= \argmin_{\substack{\bm{\phi} }}\; 
\mathbb{E}_{\bm{z}\sim p_{\bm{z}}} ||E_{\bm{\phi}}(G(\bm{z})) - \bm{z}||_2.
\end{aligned}
\label{eq:encoder}
\end{equation}
In our experiments, a ResNet \cite{he2016deep} architecture with 50 layers is used to approximate $E$.
In conclusion, the reconstructor, as a third player in GANs, guides the generator network to avoid mode-collapse and generate a wide variety of realistic images.

\subsection{Algorithm}
\label{subsec:proposed_alg}
The optimization problem in Eqn. (\ref{eq:SP_GAN0}) is not jointly convex in variables of $G$ and $D$. 
Therefore, we adopt an alternating stochastic gradient descent approach to minimize the loss functions in Eqn. (\ref{eq:SP_GAN1}), and a block coordinate descent strategy for satisfying its constraint (updating $E^*$). This procedure leads to Algorithm \ref{alg}, where the details of the updating procedure are discussed below:

\textit{Lines 1-2:}
We randomly initialize the generator and the discriminator parameters. The dictionary $\bm{\Omega}$ is trained using the online dictionary learning approach \cite{mairal2009online}.

\textit{Lines 4-9:}
We randomly select a subset of images and noise vectors. We uniformly sample a real number $\epsilon$. The random noise vector $\bm{z}_i$ is fed to the generator network and we calculate the discriminator loss in (line 8) and update the discriminator parameters (line 9).

\textit{Lines 10-13:}
We randomly select noise vectors and calculate the generator loss in (line 11). We update the generator parameters in (line 13).

\textit{Lines 14-19:}
We train the encoder network (line 15) by generating samples and solving Eqn. \eqref{eq:encoder} through a standard back propagation scheme. We skip some details due to lack of space.
Then, we randomly select noise vectors associated with the real images, and we calculate the reconstructor loss in (line 18). We update the generator parameters in (line 19).

%% file: Results.tex
\begin{figure*}[!t]
    \centering
    \begin{subfigure}[b]{1\textwidth}
        \centering
        \includegraphics[scale=0.9]{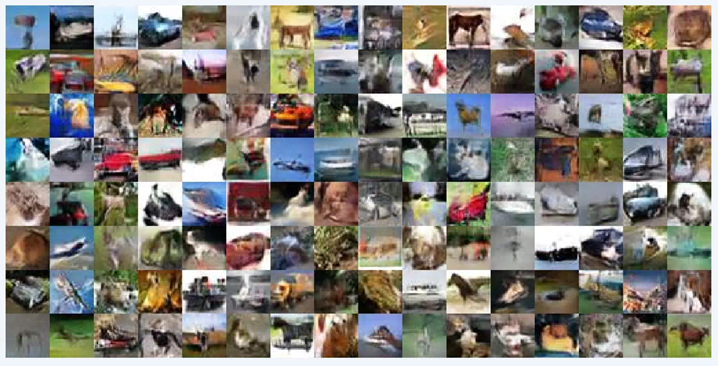}
        \caption{SPGAN (using Improved WGAN)}
    \end{subfigure}\\
    \begin{subfigure}[b]{1\textwidth}
        \centering
        \includegraphics[scale=0.476]{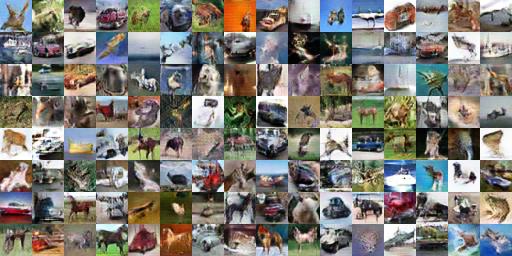}
        \caption{WGAN}
    \end{subfigure}\\
    \begin{subfigure}[b]{1\textwidth}
        \centering
        \includegraphics[scale=0.476]{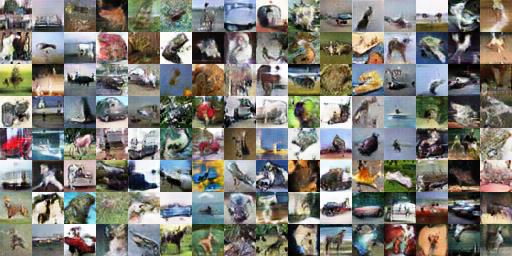}
        \caption{Improved WGAN}
    \end{subfigure}
	\caption{Generated images using CIFAR10 dataset}
	\label{fig:cifar10}
\end{figure*}

\begin{figure*}[!t]
    \centering
    \begin{subfigure}[b]{0.45\textwidth}
        \centering
        \includegraphics[scale=0.5]{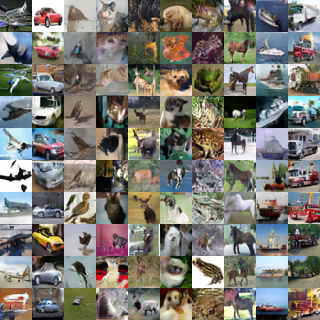}
        \caption{SPGAN (using Improved WGAN)}
    \end{subfigure}
    \begin{subfigure}[b]{0.45\textwidth}
        \centering
        \includegraphics[scale=0.5]{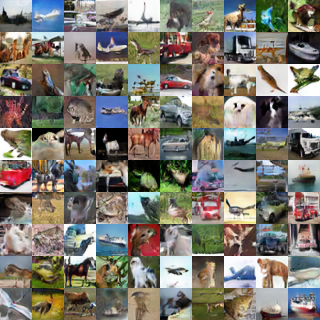}
        \caption{Improved WGAN}
    \end{subfigure}
	\caption{Generated images using CIFAR10 dataset using Resnet blocks}
	\label{fig:cifar10_res}
\end{figure*}

\begin{figure*}[!t]
    \centering
    \begin{subfigure}[b]{1\textwidth}
        \centering
        \includegraphics[scale=0.78]{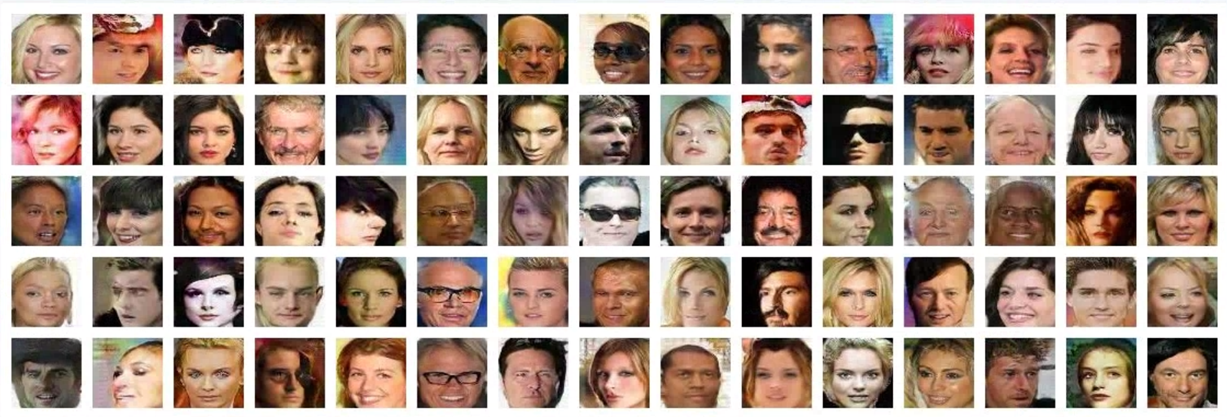}
        \caption{SPGAN (using Improved WGAN)}
    \end{subfigure}\\
    \begin{subfigure}[b]{1\textwidth}
        \centering
        \includegraphics[scale=0.32]{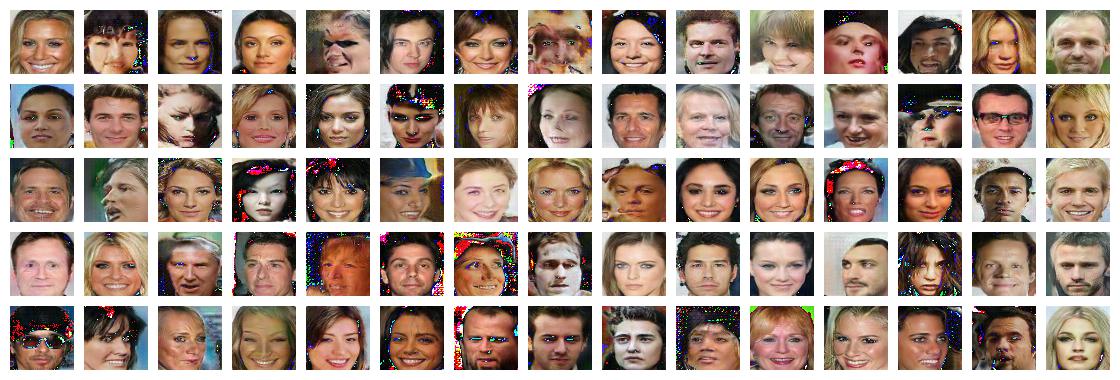}
        \caption{Improved WGAN}
    \end{subfigure}
	\caption{Generated images using celebraty face dataset}
	\label{fig:celeb}
\end{figure*}

\section{Experiments}
\label{sec:Results}
We evaluate our proposed methodology on CelebA dataset \cite{liu2015faceattributes}, and CIFAR10 object images. 
The performance of our proposed method is compared to the state-of-the-art methods. In all the following experiments, the reconstructor network $E$ is a 50-layer ResNet \cite{he2016deep} network trained on 50,000 randomly generated samples. 

\subsection{CIFAR 10 dataset}
There are 60,000 color images in CIFAR-10 dataset which are divided into 50,000 training images and 10,000 test images. The size of images in this dataset is 32$\times$32 and there are in total 10 classes in this dataset.
In order to generate a 32$\times$32 image from the 128 dimensional noise vector, a fully connected layer first expands the input dimension to a 4$\times$4$\times$1024 tensor (reshaped as tensor) and subsequently a series of three fractionally-strided convolutions transform the spatial dimension to $32\times32$ (doubling the spatial dimension in each layer). The number of channels has changed from 1024 to 512 and 100. The size of the pre-trained dictionary, $\Omega$, in this experiment is $27\times100$.

Fig. \ref{fig:cifar10} presents the images generated in this experiment and compares them with the images generated by WGAN \cite{arjovsky2017wasserstein} and Improved WGAN \cite{gulrajani2017improved}. As seen, the images generated by the proposed method have higher quality and variance compared to the images generated by the other methods. It is worth mentioning that the experiments in this section are conducted in an unsupervised setting. 
Fig. \ref{fig:cifar10_res} presents the images generated in this experiment using residual blocks in each layer of the generator network and compares them with the images generated by Improved WGAN \cite{gulrajani2017improved}. We observe that our method leads to a higher quality and diversity. 

Table \ref{tbl:1} and Table \ref{tbl:2} compare the inception score of different methods in case of a simple DCGAN generator and a generator with residual blocks respectively. Our proposed method increases the inception of WGAN and Improved WGAN in both cases. In case of a simple DCGAN generator, the inception score increased of Improved WGAN increased dramatically by 0.78 point and in case of a generator with residual blocks, the score increased by 0.09 point. 

\begin{table*}[t]
    \centering
    \begin{minipage}[t]{0.48\textwidth}
        \caption{Inception score on CIFAR10 images without residual blocks in generator}
        \centering
        \begin{tabular}{|c || c || c || c |  } 
        \hline
        \footnotesize{Method} &   
        \footnotesize{score} &
        \footnotesize{\begin{tabular}{@{}c@{}} SPGAN \\ \end{tabular}} &
        \footnotesize{\begin{tabular}{@{}c@{}} SPGAN \\ recon.\end{tabular}}\\ [0.03ex] 
        \hline\hline
        \footnotesize{ALI \cite{dumoulin2016adversarially}} & 5.36 & - & -\\
        \footnotesize{BEGAN \cite{berthelot2017began}} & 5.62 & - & -\\
        \footnotesize{WGAN \cite{arjovsky2017wasserstein}} & 5.76 & 6.1 & 6.6\\
        \footnotesize{Im-WGAN \cite{gulrajani2017improved}} &  5.92 & 6.2 & 6.7\\[0.1ex] 
        \hline
        \end{tabular}
        \label{tbl:1}
    \end{minipage}~~
    \begin{minipage}[t]{0.48\textwidth}
        \caption{Inception score on CIFAR10 images with residual blocks in generator}
        \centering
        \begin{tabular}{|c || c || c || c |  } 
        \hline
        \footnotesize{Method} &   
        \footnotesize{score} &
        \footnotesize{\begin{tabular}{@{}c@{}} SPGAN \\ \end{tabular}} &
        \footnotesize{\begin{tabular}{@{}c@{}} SPGAN \\ recon.\end{tabular}}\\ [0.03ex] 
        \hline\hline
        \footnotesize{ALI \cite{dumoulin2016adversarially}} & 5.36 & - & -\\
        \footnotesize{BEGAN \cite{berthelot2017began}} & 5.62 & - & -\\
        \footnotesize{WGAN \cite{arjovsky2017wasserstein}} & 7.73 & 7.85 & 7.88\\
        \footnotesize{Im-WGAN \cite{gulrajani2017improved}} &  7.86 & 7.93 & 7.95\\[0.1ex] 
        \hline
        \end{tabular}
        \label{tbl:2}
    \end{minipage}
\end{table*}

\subsection{CelebA dataset}
There are 200,000 color images of celebrity faces in CelebA dataset \cite{liu2015faceattributes}. The size of images in this dataset is 64$\times$64.
In order to generate a 64$\times$64 image from the 256 dimensional noise vector, at the first layer, a fully connected layer expands the input dimension to a 4$\times$4$\times$2048 tensor (reshaped as tensor) and subsequently a series of four fractionally-strided convolutions transform the spatial dimension to $64\times64$ (doubling the spatial dimension in each layer). The number of channels has changed from 2048, to1024 at the third layer, 512 at the fourth layer and to 100 at the fifth layer. The size of the pre-trained dictionary $\Omega$ in this experiment is $27\times100$.

Fig. \ref{fig:celeb} shows the generated images in this experiment and compares them with the generated images by WGAN \cite{arjovsky2017wasserstein} and Improved WGAN \cite{gulrajani2017improved}. As one can see the images generated by the proposed method have higher quality and variance compared to the images generated by the other methods.

%% file: conclusions.tex
\section{Conclusions} 
\label{sec:conclusions}
In this paper, we developed an enhanced GAN architecture, which generates images from patches, obtained through a UoS model. We also proposed a third player, called reconstructor to ensure high variability of the output images. We used two image datasets to evaluate the performance of the proposed generative adversarial method and demonstrated the remarkable advantage of regularizing the generative network by a reconstructor network and learning image characteristics at multiple scales.
The evaluation results show the merit of the proposed method for generating images. The idea can be generalized to an arbitrary number of layers in different datasets.